\documentclass{article}

     \usepackage[nonatbib,final]{neurips_2019}

\usepackage[utf8]{inputenc} 
\usepackage[T1]{fontenc}    
\usepackage{hyperref}       
\usepackage{url}            
\usepackage{booktabs}       
\usepackage{amsfonts}       
\usepackage{nicefrac}       
\usepackage{microtype}      
\usepackage{xcolor}
\usepackage{amsmath}
\usepackage{graphicx}

\title{Conditional Invertible Flow for Point Cloud Generation}

\author{%
   Michał Stypułkowski $^{12\dagger}$, Maciej Zamorski $^{13}$, Maciej Zięba $^{13}$, Jan Chorowski $^{24}$\\
   $^1$ Tooploox\\
   $^2$ University of Wrocław, Poland \\
    $^3$ Wrocław University of Science and Technology, Poland\\
    $^4$ NavAlgo\\
    $^\dagger$\texttt{michal.stypulkowski@tooploox.com}\\
}

\begin{document}

\maketitle

\begin{abstract}
 This paper focuses on a novel generative approach for 3D point clouds that makes use of invertible flow-based models. The main idea of the method is to treat a point cloud as a probability density in 3D space that is modeled using a cloud-specific neural network. To capture the similarities between point clouds we rely on parameter sharing among networks, with each cloud having only a small embedding vector that defines it. 
We use invertible flows networks to generate the individual point clouds, and to regularize the embedding space. We evaluate the generative capabilities of the model both in qualitative and quantitative manner. 

\end{abstract}

\section{Introduction}

In this paper we propose a novel method for generating 3D point clouds that utilize the flow-based models. Conceptually, we treat each point in 3D space as an instance in data space and we represent a set of points that constitutes the point cloud using a flow-based neural network. However, to capture the similarities between point clouds, each of the networks shares the parameters and each point cloud is described with an individual embedding that is used as conditional factor in the flow-based model. In order to generate the point cloud we sample an embedding from the assumed prior distribution, then we sample points from the flow assuming given embedding value. The prior distribution for the embedding space is achieved with an additional embedding-level flow.

The proposed model can be interpreted as a meta-learner, where each point cloud is treated as a separate training set that is used to train a single network. However, instead of retraining a model for each point cloud we use an additional embedding space that controls the shape of the point cloud. Harnessing meta-learning techniques enable us to discover the embedding for new point cloud with gradient-based approach by keeping the parameters of flow-based model unchanged. Because of the point-level generative procedure our model is order-invariant and insensitive to the different numbers of points among point clouds. 

We identify the following contribution of our work: (i) a novel generative model for 3D point clouds, coherent in terms of mathematics thanks to direct likelihood optimization and change of variable formula, (ii) point-level generative approach that controls the sampling process in every step, and is insensitive to the ordering and different numbers of the points, (iii) a new meta-learning approach, where each point cloud (as separate training subset) is represented by unique embedding that can be parameterized.  

\section{Model description}

\begin{figure}
  \centering
  \begin{tabular}{p{.49\textwidth}p{.49\textwidth}}
  \centering
  \vspace{1cm}\includegraphics[width=0.45\textwidth]{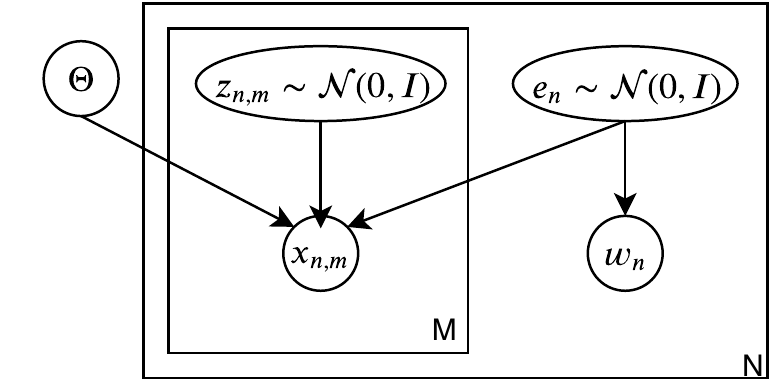} &
  \vspace{0pt}\includegraphics[width=0.45\textwidth]{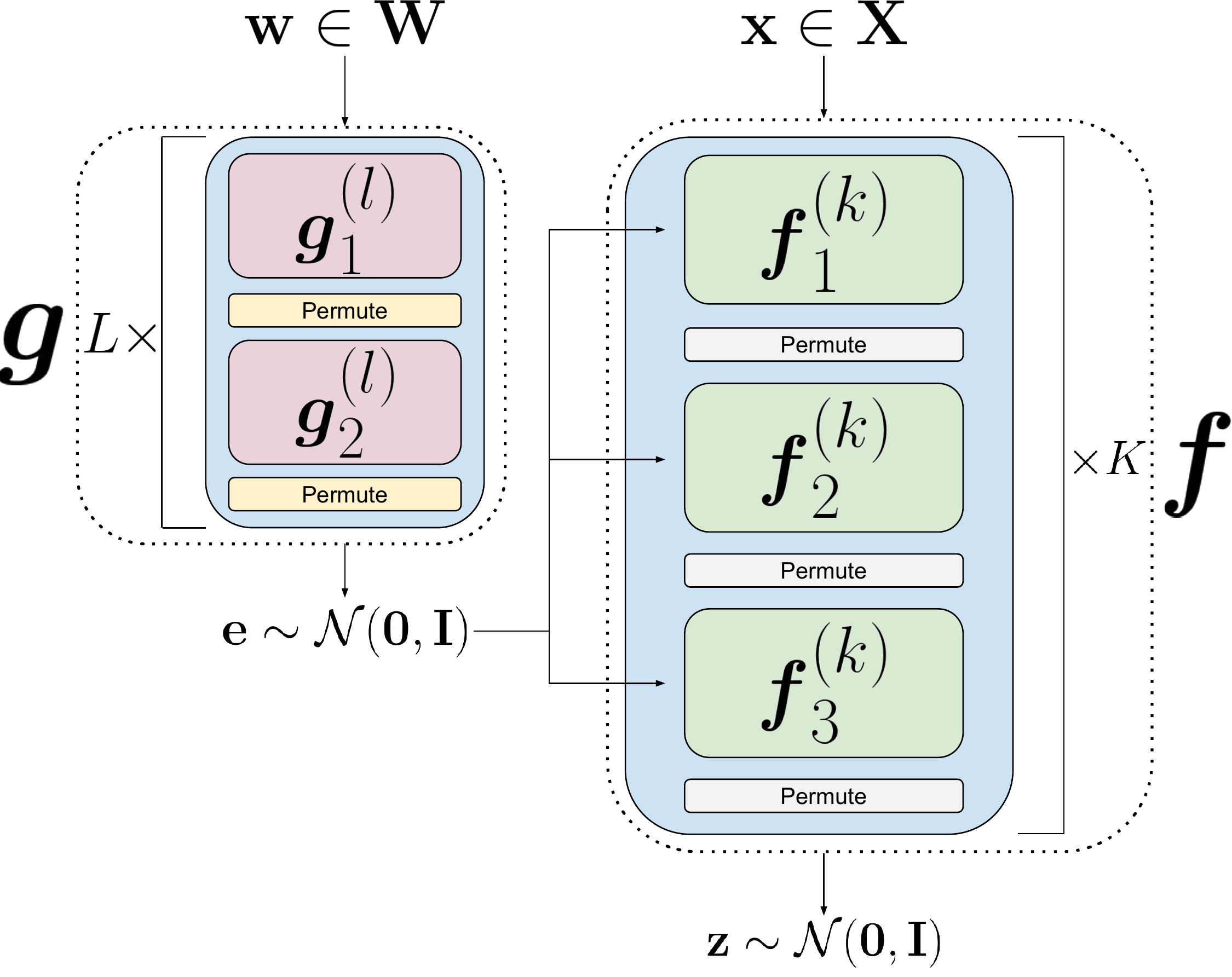} \\
  \centering a) & \centering b)
  \end{tabular}
  \caption{a) Our model generates individual 3D points $\mathbf{x}_{n,m}$ conditioned on a point-cloud embeddings $\mathbf{e}_n$. To regularize the embedding space we introduce additional observed variables $\mathbf{w}_n$ set to the location of the point cloud given by MDS. b) We implement our model using invertible normalizing flows $\mathbf{f}$ and $\mathbf{g}$.}
  \label{fig:model}
\end{figure}

Our model is composed of two invertible normalizing flows, $\mathbf{f}$ and $\mathbf{g}$ (see Figure \ref{fig:model}). The model $\mathbf{f}$ is responsible for transforming single-point data space $X$ to the normally distributed $Z$ space. The model $\mathbf{f}$ is conditioned by the latent factor $\mathbf{e}$ that can be seen as embedding representation of the point cloud. The role of $\mathbf{g}$ is to map some observable point cloud representation $\mathbf{w}$ to the normally distributed latent space $E$ to achieve latent point cloud representation from a given prior distribution.

For generative purposes, we first sample point cloud embedding $\mathbf{e}$ from the assumed prior. Next, for the sampled $\mathbf{e}$ we utilize model $\mathbf{f}$ to generate a point cloud by sampling desired number of points $\mathbf{z}$ from $Z$ space and using invertible transformation $\mathbf{f}^{-1}(\mathbf{z}, \mathbf{e})$ to generate points from the data space $X$.

\subsection{Flow-based model} \label{sec:flow_model}
The main idea of flow-based models is to find invertible transformation between some simple distribution $p_Z$ and complex one $p_X$. We assume that there exists a bijection $\mathbf{f}$ such that $\mathbf{z}=\mathbf{f}(\mathbf{x})$, where $\mathbf{z}\sim p_Z$ and $\mathbf{x}\sim p_X$. We make use of the change of variables formula for density functions
\begin{align}
    p_X(\mathbf{x}) = p_Z(\mathbf{f}(\mathbf{x})) \Big{|} \det \frac{\partial \mathbf{f}(\mathbf{x})}{\partial \mathbf{x}^T} \Big{|},
\end{align}
where $\Big{|} \det \frac{\partial \mathbf{f}(\mathbf{x})}{\partial \mathbf{x}^T} \Big{|}$ is the absolute value of the determinant of the Jacobian of the transformation $\mathbf{f}$ at point $\mathbf{x}$. With proper definition of $\mathbf{f}$ we are able to train model with direct likelihood of the data. We follow \cite{dinh2016density} in construction of our flows and define $\mathbf{f}$ as combinations of simple flows $\mathbf{f}^{(k)}_i$ each followed by permutation of the dimensions, where $\mathbf{f}^{(k)}_i$ is defined as:
\begin{align}
    &\mathbf{h}_{1:d} = \mathbf{x}_{1:d} \label{simple_flow}\\ 
    &\mathbf{h}_{d+1:D} = \mathbf{x}_{d+1:D} \odot \exp (M(\mathbf{x}_{1:d})) + A(\mathbf{x}_{1:d}), \nonumber
\end{align}
where $D$ is dimension of $\mathbf{x}$ and $d<D$. This transformation is easily invertible and the logarithm of the Jacobian equals $\sum_i M(\mathbf{x}_{1:d})_i$. Note that there are no constraints on either $M$ or $A$ thus we choose them to be neural networks.

\subsection{Conditional Flow}
Instead of treating $\mathbf{x}$ as a whole point cloud, we take $p_X$ as a distribution of points over the surface of selected point cloud. This specified point cloud is represented by its embedding $\mathbf{e}$. If we add $\mathbf{e}$ as a second argument to $\mathbf{f}$, we are able to choose the distribution from which we want to sample points.

We introduce second flow $\mathbf{g}$ which is responsible for mapping vectors $\mathbf{w}$ into vectors $\mathbf{e}\sim \mathcal{N}(\mathbf{0},\mathbf{I})$. $\mathbf{w}$ is observable representation of single point cloud. It is obtained using Multidimensional Scaling algorithm, which transforms matrix of Chamfer Distances between every pair of point clouds in training set into lower dimensional space $W$. 

Given the data point $\mathbf{x}$ and the corresponding vector $\mathbf{w}$, our main goal is to find the joint distribution $p_{X,W}$. Using the same approach as in the single-flow model and the fact that $Z$ and $E$ are independent, we get
\begin{align}
    p_{X,W}(\mathbf{x},\mathbf{w}) &= p_{Z,E} (\mathbf{f}(\mathbf{x},\mathbf{e}),\mathbf{g}(\mathbf{w})) \cdot \Bigg{|} \det 
        \begin{pmatrix} 
            \frac{\partial \mathbf{f}(\mathbf{x},\mathbf{e})}{\partial \mathbf{x}^T} & \frac{\partial \mathbf{f}(\mathbf{x},\mathbf{e})}{\partial \mathbf{w}^T} \\
            0 &  \frac{\partial \mathbf{g}(\mathbf{w})}{\partial \mathbf{w}^T} 
        \end{pmatrix} 
    \Bigg{|} \\
    &=p_Z(\mathbf{f}(\mathbf{x},\mathbf{e})) \cdot p_E(\mathbf{g}(\mathbf{w})) \cdot \Big{|} \det \frac{\partial \mathbf{f}(\mathbf{x},\mathbf{e})}{\partial \mathbf{x}^T}\Big{|} \cdot \Big{|} \det \frac{\partial \mathbf{g}(\mathbf{w})}{\partial \mathbf{w}^T}\Big{|}, \nonumber
\end{align}
where $\mathbf{z} = \mathbf{f}(\mathbf{x},\mathbf{e})$ and $\mathbf{e}=\mathbf{g}(\mathbf{w})$.

The dataset is composed of set of pairs $(\mathbf{x}_{n,m}, \mathbf{w}_n)$, where $\mathbf{x}_{n,m}$ is the $m$-th point from the $n$-th point cloud and $\mathbf{w}_{n}$ is the vector representing the $n$-th point cloud. We are able to train our model optimizing log-likelihood of the data and the vectors $\mathbf{w}$
\begin{align}
    \sum_n \sum_m \log \big{(} p_{X,W}(\mathbf{x}_{n,m},\mathbf{w}_n) \big{)} = &\sum_n \sum_m \bigg{[}\log \big{(}p_Z(\mathbf{f}(\mathbf{x}_{n,m},\mathbf{e}_{n})) \big{)} + \log \big{(}p_E(\mathbf{g}(\mathbf{w}_{n}))\big{)} \label{eq:loss} \\ 
    &+ \log \Big{|} \det \frac{\partial \mathbf{f}(\mathbf{x}_{n,m},\mathbf{e}_n)}{\partial \mathbf{x}^T}\Big{|}
    + \log \Big{|} \det \frac{\partial \mathbf{g}(\mathbf{w}_n)}{\partial \mathbf{w}^T}\Big{|}\bigg{]}. \nonumber
\end{align}

In practice, embedding $\mathbf{e}$ as a second argument of $\mathbf{f}$ is passed to both $M$ and $A$ in every simple flow $\mathbf{f}^{k}_{i}$. Due to the fact that $\mathbf{x}$ is $3$-dimensional vector, we are able to transform one dimension in every simple flow $\mathbf{f}^{k}_{i}$, i.e. $d$ in Eq. \ref{simple_flow} is equal to $2$. We shift data dimensions one place right after each simple flow to transform all of the dimensions.
Inverse function $\mathbf{f}^{-1}$ is defined as inverse function of $\mathbf{f}$ with respect to its first argument, i.e. $\mathbf{z} = \mathbf{f}(\mathbf{x},\mathbf{e})$ and $\mathbf{x} = \mathbf{f}^{-1}(\mathbf{z},\mathbf{e})$.
We define $\mathbf{g}$ analogously as in \ref{sec:flow_model}. We split dimensions in half and swap these halves after each $\mathbf{g}^{l}_{j}$.

\section{Related work}

The application of deep generative models to 3D shapes were initially studied in \cite{wu20153d} and \cite{wu2016learning}. 
The former work focuses on application of Convolutional Deep Belief Network on 3D point grids. 
The later uses an adversarial training to learn the underlying 3D voxels distribution. In \cite{achlioptas2017learning} authors provide a two stage approach with simple auto-encoder and GAN model trained in stacked mode. Authors of \cite{zamorskiadversarial} provide interesting variation of adversarial autoencoder for generating 3D point clouds. In \cite{li2018point} the end-to-end GAN for point clouds is provided. One of the recent papers \cite{yang2019pointflow} utilizes VAE together with continuous normalizing flows to generate 3D point clouds.

In contrary to the reference approaches our model operates on single points rather than on entire point clouds, delivers coherent likelihood optimization rather then ELBO and is invariant to the number of the points stored in the point cloud.

\section{Experiments}
The goal of the experiments is to provide quantitative and qualitative analysis of generative capabilities of the model in terms of the quality of generated samples, interpolation and coverage and minimum matching distance metrics. For all experiments we used point clouds from ShapeNet dataset limited to objects representing chairs and we split the data to train and test sets with $90\%$-$10\%$ proportions. 

\subsection{Architecture}
We implemented both $\mathbf{f}$ and $\mathbf{g}$ as normalizing flows. $\mathbf{f}$ is a combination of $10$ segments $\mathbf{f}^{(k)}$ for $k=1,...,10$ with $3$ blocks $\mathbf{f}^{(k)}_i$ for $i=1,2,3$ each. Every block is defined by two ResNets. Analogically $\mathbf{g}$ has $5$ segments with $2$ blocks each. We chose dimensions of $\mathbf{w}$ and $\mathbf{e}$ to equal $64$. Priors $p_Z$ and $p_E$ are $d$-dimensional standard normal distributions, where $d$ equals to $3$ and $64$ respectively. We used Adam optimizer with default parameters and learning rate $10^{-4}$ decaying every $40$ epochs by factor $0.8$.

\begin{figure}
  \centering
  \includegraphics[width=1\textwidth]{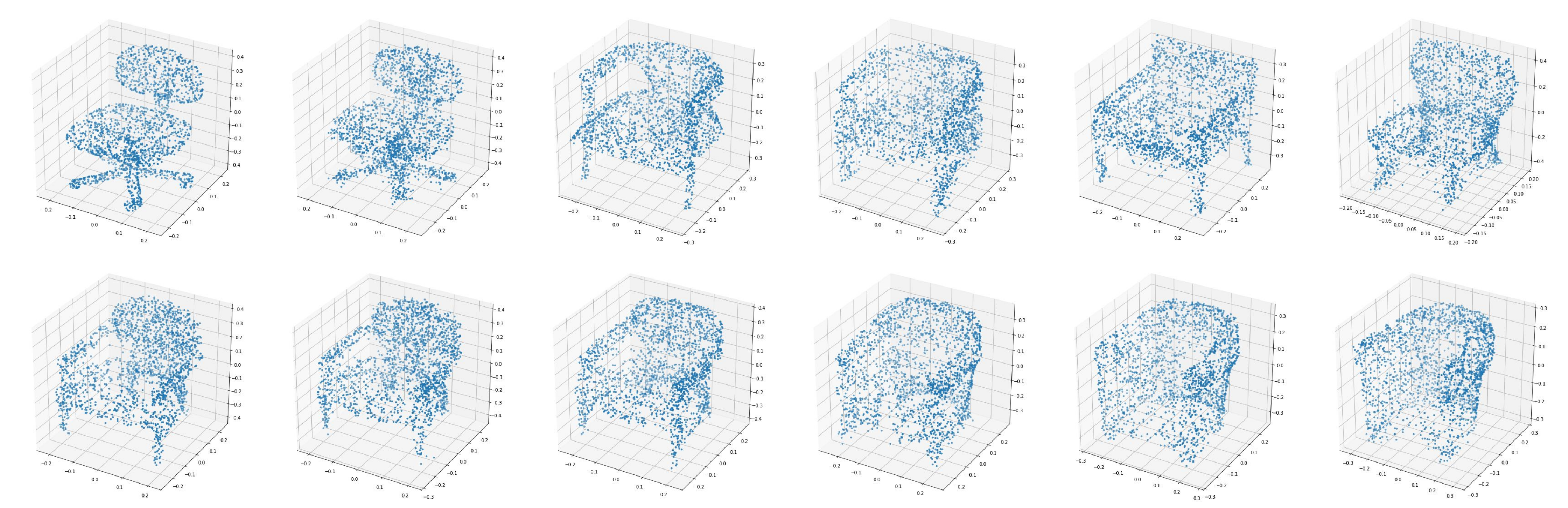}
  \caption{Top row (from the left): chair from the training set and its reconstruction, chair from the test set and its reconstruction, two samples generated randomly from prior distributions. Bottom row: Interpolation between two reconstructed point clouds.}
  \label{fig:recon_train}
\end{figure}

\begin{table}
  \caption{MMD and Coverage results compared to reference models}
  \label{tab:mmd_cov}
  \centering
  \begin{tabular}{c|ccc}
    \toprule
    Model & S-CIF & CIF & l-GAN \cite{achlioptas2017learning}\\
    \cmidrule(r){1-4}
     COV-CD & $29.79\%$ & $48.23\%$ & $59.4\%$\\
    MMD-CD &  $0.0025$ & $0.0018$ & $0.0020$ \\
    \bottomrule
  \end{tabular}
\end{table}

\subsubsection{Results}
Qualitative results of reconstruction, sampling and interpolation are shown on Figure \ref{fig:recon_train}.

\textbf{Reconstruction}
Reconstructions of training set point cloud are made with known embeddings $\mathbf{e}$ (or equivalently $\mathbf{g(w)}$) used during training. This experiment shows $\mathbf{f}$'s ability of sampling points from desired distribution. Generated point clouds are sharp and they are good reconstructions of the training set.

Reconstruction of the test set object requires finding the proper vector $\mathbf{e}$ in the embedding space $E$. In order to do that, we initialize $\mathbf{e} \sim p_E$ and update it by maximizing the likelihood 
\begin{align}
     p_{X}(\mathbf{x};\mathbf{e}) =p_Z(\mathbf{f}(\mathbf{x};\mathbf{e})) \cdot \Big{|} \det \frac{\partial \mathbf{f}(\mathbf{x};\mathbf{e})}{\partial \mathbf{x}^T}\Big{|}.
\end{align}
We perform optimization only with respect to $\mathbf{e}$, having $\mathbf{f}$'s weights frozen. Quality of the reconstruction varies based on the uniqueness of the object. Best results are achieved on simpler shapes but on more complicated ones results are still satisfying.

\textbf{Sampling}
To sample a new point cloud, we first sample embedding $\mathbf{e}$ from prior distribution $p_E$ to obtain the conditioning term for $\mathbf{f}$. Next we sample desired number of points $\mathbf{z}$ from $p_Z$. We pass $\mathbf{e}$ and $\mathbf{z}$ samples to $\mathbf{f}^{-1}$ in order to generate various shapes of chairs.

\textbf{Interpolation}
We performed interpolation in embedding space $E$ between two clouds from training set. Results show that embeddings are arranged in natural and logical way. Transitions between each sample is smooth and there are no huge visual differences.

\textbf{Quantitative} We follow evaluations made in \cite{achlioptas2017learning} to check overall quality of samples and its generalization. We calculated Minimum Matching Distance (MMD) and Coverage (COV) of sampled point clouds with respect to the test set using Chamfer Distance (CD). We report the results in Table \ref{tab:mmd_cov} comparing our Conditional Invertible Flow (CIF) model with Single Conditional Invertible Flow (S-CIF) model that uses simple MLP instead of $\mathbf{g}$ flow and KL divergence to enforce $\mathbf{e}$ to be normally distributed and with l-GAN \cite{achlioptas2017learning} trained with CD. Our approach achieves the best results in terms of MMD but l-GAN outperforms our approach with respect to COV. However, no specific model selection was performed with respect to our method on current stage of the development.

\section{Conclusion and future work}
In this paper we provide the novel point-level order-invariant generative model for 3D point clouds that is coherent in terms of mathematics thanks to direct likelihood optimization. We provide some quantitative and qualitative results that are promising for the future works. In next steps we are going to stabilize the training between two flow models, search for better flow-based architectures and make the deeper qualitative analysis with respect to log-likelihood values.  

\subsubsection*{Acknowledgments}
We thank the PLGrid project for computational resources on the Prometheus cluster. We also thank Tomasz Konopczyński for technical support and valuable advices.


\begin{thebibliography}{1}

\bibitem{achlioptas2017learning}
P.~Achlioptas, O.~Diamanti, I.~Mitliagkas, and L.~Guibas.
\newblock Learning representations and generative models for 3d point clouds.
\newblock {\em arXiv preprint arXiv:1707.02392}, 2017.

\bibitem{dinh2016density}
L.~Dinh, J.~Sohl-Dickstein, and S.~Bengio.
\newblock Density estimation using real nvp.
\newblock {\em arXiv preprint arXiv:1605.08803}, 2016.

\bibitem{li2018point}
C.-L. Li, M.~Zaheer, Y.~Zhang, B.~Poczos, and R.~Salakhutdinov.
\newblock Point cloud gan.
\newblock {\em arXiv preprint arXiv:1810.05795}, 2018.

\bibitem{wu2016learning}
J.~Wu, C.~Zhang, T.~Xue, B.~Freeman, and J.~Tenenbaum.
\newblock Learning a probabilistic latent space of object shapes via 3d
  generative-adversarial modeling.
\newblock In {\em Advances in neural information processing systems}, pages
  82--90, 2016.

\bibitem{wu20153d}
Z.~Wu, S.~Song, A.~Khosla, F.~Yu, L.~Zhang, X.~Tang, and J.~Xiao.
\newblock 3d shapenets: A deep representation for volumetric shapes.
\newblock In {\em Proceedings of the IEEE conference on computer vision and
  pattern recognition}, pages 1912--1920, 2015.

\bibitem{yang2019pointflow}
G.~Yang, X.~Huang, Z.~Hao, M.-Y. Liu, S.~Belongie, and B.~Hariharan.
\newblock Pointflow: 3d point cloud generation with continuous normalizing
  flows.
\newblock {\em arXiv preprint arXiv:1906.12320}, 2019.

\bibitem{zamorskiadversarial}
M.~Zamorski, M.~Zieba, P.~Klukowski, R.~Nowak, K.~Kurach, W.~Stokowiec, and
  T.~Trzcinski.
\newblock Adversarial autoencoders for compact representations of 3d point
  clouds.
\newblock {\em arXiv preprint arXiv:1811.07605}, 2018.

\end{thebibliography}


\end{document}